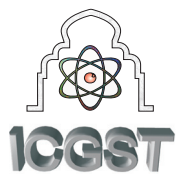

www.icgst.com

# Face Detection from still and Video Images using Unsupervised Cellular Automata with K means clustering algorithm

P.Kiran Sree [1], I Ramesh Babu [2]

1. Associate Professor, Department of C.S.E, S.R.K.I.T,Vijayawada, profkiran@yahoo.com.
2. Head of the Department & Professor, C.S.E,Acharya Nagarjuna University, Guntur

## Abstract
Pattern recognition problem rely upon the features inherent in the pattern of images. Face detection and recognition is one of the challenging research areas in the field of computer vision. In this paper, we present a method to identify skin pixels from still and video images using skin color. Face regions are identified from this skin pixel region. Facial features such as eyes, nose and mouth are then located. Faces are recognized from color images using an RBF based neural network. Unsupervised Cellular Automata with K means clustering algorithm is used to locate different facial elements. Orientation is corrected by using eyes. Parameters like inter eye distance, nose length, mouth position, Discrete Cosine Transform (DCT) coefficients etc. are computed and used for a Radial Basis Function (RBF) based neural network. This approach reliably works for face sequence with orientation in head, expressions etc.

## 1. Introduction
The technological advancement in the area of digital processing and imaging has led to the development of different algorithms for various applications such as automated access control, surveillance etc. For automated access control, most common and accepted method is based on face detection and recognition. Face recognition is one of the active research areas with wide range of applications. The problem is to identify facial image/region from a picture/image. Generally pattern recognition problems rely upon the features inherent in the pattern for efficient solution. Though face exhibits distinct features which can be recognized almost instantly by  human eyes, it is very difficult to extract these features and use these features by a computer. Human can identify faces even from a caricature. The challenges associated with face detection and recognition are pose, occlusion, skin color, expression, presence or absence of structural components, effects of light, orientation, scale, imaging conditions etc. Most of the currently proposed methods use parameters extracted from facial images.  For access control application, the objective is to authenticate a person based on the presence of a recognized face in the database.

In this paper, skin and non-skin pixels are separated and the pixels in the identified skin region are grouped to obtain face region. From the detected face area, the facial features such as eyes, nose and mouth are located. Rest of the paper is organized as follows. Section 2 gives background and related works. Section 3 discusses the proposed method. Results are given in section 4. Conclusions and future works are given in section 5.

## 2. Background and related work
 A lot of research has been going on in the area of human face detection and recognition [3]. Most face detection and recognition methods fall into two categories: Feature based and Holistic. In feature-based method, face recognition relies on localization and detection of facial features such as eyes, nose, mouth and their geometrical relationships. In holistic approach, entire facial image is encoded into a point on high dimensional space.  Principal Component Analysis (PCA) and Active Appearance Model (AAM) [9] for recognizing faces are based on holistic approaches. In another approach, fast and accurate face detection is performed by skin color learning by neural network and segmentation technique [4]. Independent Component Analysis (ICA) was performed on face images under two different conditions [8]. In one condition, image is treated as a random variable and pixels are treated as outcomes and in the second condition pixels are treated as random variables and image as outcome. Facial expressions are extracted from the detailed analysis of eye region images is given in [2].  Large range of human facial behavior is handled by recognizing facial muscle actions that produce expressions is given in [10]. Video based face recognition is explained in [5].

## 3. Method
The method explains detection of faces from video frames and still images. This is followed by extraction of facial features.

## 3.1 Facial region identification

The first step in face detection problem is to extract facial area from the background. In our approach, both still and video images are used for face detection. Image frames from video are extracted first. The input images contain regions other than face such as hair, hat etc. Hence it is required to identify the face. Each pixel in the image is classified as skin pixel or non skin pixel. Different skin regions are detected from the image. Face regions are identified from the detected skin region as in [7] which addressed the problem of face detection in still images. Some randomly chosen frames with different head pose, far away from the camera, and expressions from video face images are extracted as shown in figure 1. The difference image at various time instances is shown in figure 2.

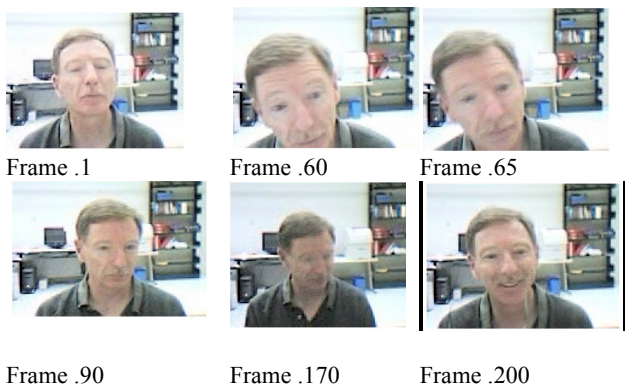

Frame .1    Frame .60    Frame .65

Frame .90    Frame .170    Frame .200

Figure 1. Some frames from face video database

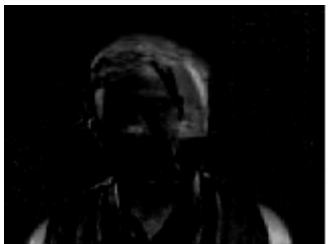

Figure 2. Difference image

The portion of the image which is moving is assumed as head. The face detection algorithm used RGB color space for the detection of skin pixels. The pixels corresponding to skin color of the input image is classified according to certain heuristic rules. The skin color is determined from RGB color space as explained in [1] and [6]. A pixel is classified as skin pixel if it satisfies the following conditions.

$R > 95$ AND $G > 40$ AND $B > 20$ AND $\max\{R, G, B\}-\min\{R, G, B\} > 15$ AND $|R-G| > 15$ AND $R > G$ AND $R > B$     (1)

OR

$R > 220$ AND $G > 210$ AND $B > 170$ AND $|R-G| \leq 15$ AND $R > B$ AND $G > B$     (2)

Edge detection is performed in each frame. Edge detection and skin regions identified from the color images of video frames and still images are shown in figure 3 and figure 4. From these skin regions, it is possible to identify whether a pixel belongs to skin region or not. To find the face regions, it is necessary to categorize the skin pixels in to different groups so that it will represent some meaningful groups such as face, hand etc. Connected component labeling is performed to classify the pixels. In the connected component labeling operation, pixels are connected together geometrically. In this, we used 8-connected component labeling so that each pixel is connected to its eight immediate neighbors. At this stage, different regions are identified and we have to classify each region as a face or not. This is done by finding the skin area of each region. If the height to width ratio of those skin region falls with in the range of golden ratio $((1+\sqrt{5})/2 \pm$ tolerance), then that region is considered as a face region.

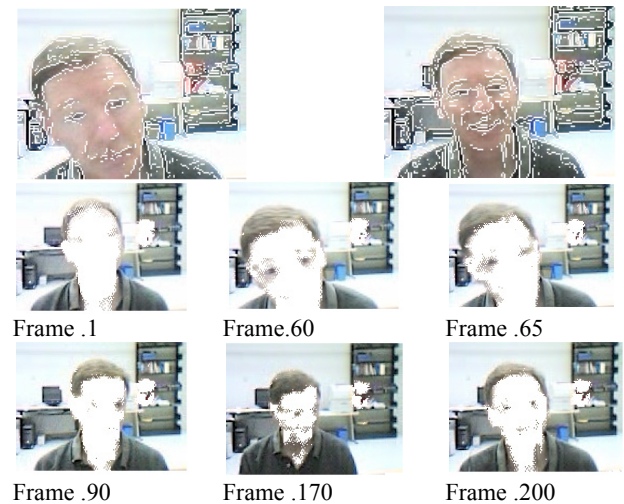

Frame .1    Frame.60    Frame .65

Frame .90    Frame .170    Frame .200

Figure 3. Edge detection and Skin region identification from video frames

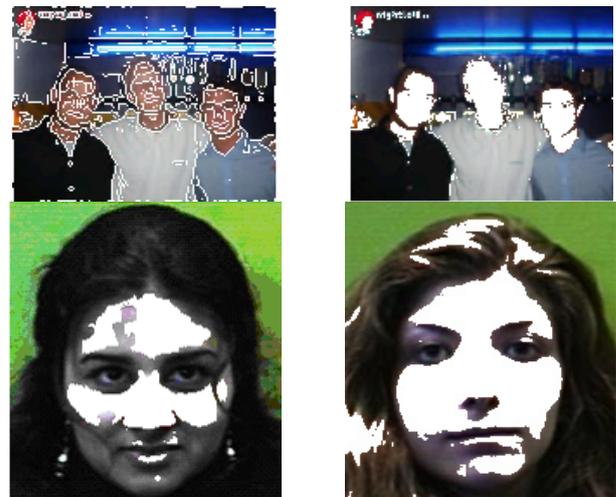

Figure 4. Edge detection and Skin region identification from still images

## 3.2. Facial feature Extraction
### 3.2.1 Edge Detection

There are many ways to perform edge detection. However, the most may be grouped into two categories, gradient and Laplacian. The gradient method detects the edges by looking for the maximum and minimum in the first derivative of the image. The Laplacian method searches for zerocrossings in the second derivative of the image to find edges. This first figure shows the edges of an image detected using the gradient method (Roberts, Prewitt, Sobel) and the Laplacian method (Marrs-Hildreth). Compared to the known methods in the literature, our algorithm has a number of advantages. It provides edge strength measures that have a straightforward geometric interpretation and supports a classification of edge points into several subtypes. We give a definition of optimal edge detectors and compare our algorithm to this theoretical model. We have carried out extensive tests using real range images acquired by four range scanners with quite different characteristics. Using a simple contour closure technique, we show that our edge detection method is able to achieve a complete range image segmentation into regions. This edge-based segmentation approach turns out to be superior to many region-based methods with regard to both segmentation quality and computational efficiency. The good results that were achieved demonstrate the practical usefulness of our edge detection algorithm.

### 3.2.3. Segmentation

Image segmentation is a long standing problem in computer vision. There are different segmentation techniques which divides spatial area with in an image to different meaningful components. Segmentation of images is based on the discontinuity and similarity properties of intensity values. Cluster analysis is a method of grouping objects of similar kind in to respective categories. It is an exploratory data analysis tool which aims at sorting different objects in to groups in a way that degree of association between two objects is maximal if they belong to same group and minimal otherwise. K-Means is one of the unsupervised learning algorithms that solve the clustering problems. The procedure follows a simple and easy way to classify a given data set through a certain number of clusters (assume Kclusters) fixed apriori. The idea is to define K centroids, one for each cluster. These centroids should be placed in a cunning way because of different location causes different results. So the better choice is to place them as much as possible far away from each other. The next step is to take each point belonging to a given dataset and associate it to the nearest centroid. When no point is pending, the first step is completed and an early groupage is done. At this point K new centroids need to be re-calculated as barycentres of the clusters resulting from previous step. After getting these K new centroids, a new binding has to be done between the same dataset points and the nearest new centroid. A loop has been generated. As a result of this loop we may notice that the K centroids change their location step by step until no more changes are done. If we know the number of meaningful groups/classes based on the range of pixel intensity, weighted K-means clustering can be used to cluster the spatial intensity values.

In facial images the skin color and useful components can be generally classified as two different classes. But if we use two class based clustering, it may result in components that may be still difficult to identify. So we used three classes and are able to cluster the data in useful manner. Initial cluster centers are calculated using histogram. Then K-means clustering algorithm computes distance between the different pixels and cluster centers and selects a minimum distance cluster for each pixel. This process continues until all pixels are classified properly.

The results of clustering algorithm are shown in figure 5. Class–I is selected since it is possible to separate the components properly compared to other classes. Then connectivity algorithm is applied to all the components in the clustered face.

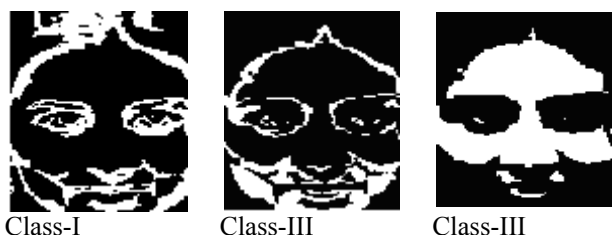

Class-I      Class-III      Class-III
Figure 5.Clustering

### 3.2.2. Feature extraction

Eye regions are located in the upper half of skin region and can be extracted using the area information of all the connected components. Using eye centers, orientation is corrected by rotation transformation. After calculating the inter eye distance, nose and mouth are identified as they generally appear along the middle of two eyes in the lower half. The inter eye distance, nose length and mouth area are computed

The DCT is employed to extract the input features to build a face recognition system, and the flexible neural tree is used to identify the faces. Based on the pre-defined instruction/operator sets, a flexible neural tree model can be created and evolved. This framework allows input features selection, over-layer connections and different activation functions for the various nodes involved. Empirical results indicate that the proposed framework is efficient for face recognition. First, in this paper, the dimensionality of the original face image is reduced by using the DCT and the upper-left corner of the DCT matrix is selected to be the features of face image. Next, the proper feature are abstracted from the truncated DCT coefficient matrix by 2DLDA. The proposed algorithms are compared with both the DCT-based algorithm and the DCT+LDA algorithm which are proposed for face recognition.

To keep track of the overall information content in the face area, we used DCT. On applying DCT, most of the energy values can be represented by a few coefficients.

First 64 x 64 coefficients are taken as part of the feature set. The face area calculated from the skin region is taken as another parameter.

### 3.3. Radial basis function

Radial functions are a special class of function. A general and efficient design approach using a radial basis function (RBF) neural classifier to cope with small training sets of high dimension, which is a problem frequently encountered in face recognition, is presented in this paper. A neural network based face recognition system is presented in this paper. The system consists of two main procedures. The first one is face features extraction using Pseudo Zernike Moments (PZM) and the second one is face classification using Radial Basis Function (RBF) neural network.

Their characteristic feature is that their response decreases (or increases) monotonically with distance from a central point. The centre, the distance scale, and the precise shape of the radial function are parameters of the model, all fixed if it is linear. A typical radial function is the Gaussian which, in the case of a scalar input, is

$$h(x) = exp(-(x-c)^2/r^2). \qquad (4)$$

Its parameters are its centre $c$ and its radius r. Universal approximation theorems show that a feed forward network with a single hidden layer with non linear units can approximate any arbitrary function[Y]. No learning is involved in RBF networks. For pattern classification problems, the numbers of input nodes are equal to the number of elements in the feature vector and the numbers of output nodes are equal to the number of different clusters. Unsupervised learning studies how systems can learn to represent particular input patterns in a way that reflects the statistical structure of the overall collection of input patterns. By contrast with supervised learning or reinforcement learning, there are no explicit target outputs or environmental evaluations associated with each input; rather the unsupervised learner brings to bear prior biases as to what aspects of the structure of the input should be captured in the output.

### 3. Unsupervised Learning
### 3.1 Machine learning, statistics, and information theory

Almost all work in unsupervised learning can be viewed in terms of learning a probabilistic model of the data. Even when the machine is given no supervision or reward, it may make sense for the machine to estimate a model that represents the probability distribution for a new input $x_t$ given previous inputs $x_1, \ldots, x_{t-1}$ (consider the obviously useful examples of stock prices, or the weather).

That is, the learner models $P(x_t|x_1, \ldots, x_{t-1})$. In simpler cases where the order in which the inputs arrive is irrelevant or unknown, the machine can build a model of the data which assumes that the data points $x_1$, $x_2$, . . . are independently and identically drawn from some distribution $P(x)^2$.

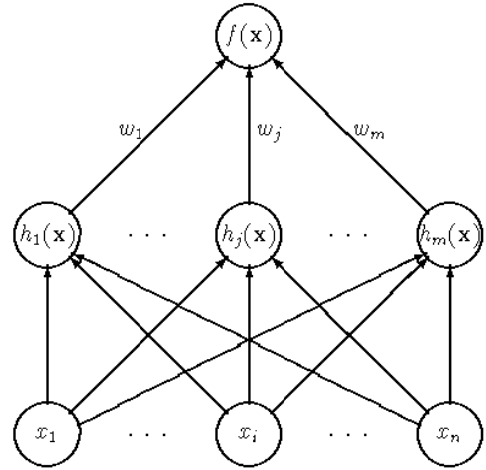

Figure 6. Radial Basis Function Network

### 3.2 A Distinct K-Means Algorithm

The k-means algorithm with the distinct difference allows for different number of clusters, while the k-means assumes that the number of clusters is known a priori. The objective of the k-means algorithm is to minimize the within cluster variability. The objective function (which is to be minimized) is the sums of squares distances of each DNA sequence and its assigned cluster center.

$$SS_{distances} = \sum_{\forall x} [x - c(x)]^2 \qquad (5)$$

where C(x) is the mean of the cluster that DNA position x is assigned to Minimizing the $SS_{distances}$ is equivalent to minimizing the Mean Squared Error (MSE). The MSE is a measure of the within cluster variability.

$$MSE = \frac{\sum_{\forall x} [x - c(x)]^2}{(N - c)b} = \frac{SS_{distances}}{(N - c)b} \qquad (6)$$

where N is the number of DNA distance centers, c indicates the number of clusters, and b is the number of spectral bands. K-means is very sensitive to initial starting values. For two classifications with different initial values and resulting different classification one could choose the classification with the smallest MSE (since this is the objective function to be minimized). However, as we show later, for two different initial values the differences in respects to the MSE are often very small while the classifications are very different. Visually it is often not clear that the classification with the smaller MSE is truly the better classification.

## 4. FMACA Based Tree-Structured Classifier

Like decision tree classifiers, FMACA based tree structured classifier uses the distinct k-means algorithm recursively partitions the training set to get nodes (attractors of a FMACA) belonging to a single class. Each node (attractor basin) of the tree is either a leaf indicating a class; or a decision (intermediate) node which specifies a test on a single FMACA, according to equations 1,2.

Suppose, we want to design a FMACA based pattern classifier to classify a training set $S = \{S1, S2, \cdot \cdot , SK\}$ into $K$ classes. First, a FMACA with $k$-attractor basins is generated. The training set $S$ is then distributed into $k$ attractor basins (nodes). Let, $S'$ be the set of elements in an attractor basin. If $S'$ belongs to only one class, then label that attractor basin for that class. Otherwise, this process is repeated recursively for each attractor basin (node) until all the examples in each attractor basin belong to one class. Tree construction is reported in [7]. The above discussions have been formalized in the following algorithm. We are using genetic algorithm classify the training set.

***Algorithm 1***: **FMACA Tree Building (using distinct K means algorithms)**
 Input    :    Training set $S = \{S1, S2, \cdot \cdot , SK\}$
 Output:    FMACA Tree.
**Partition**$(S, K)$
Step 1: Generate a FMACA with $k$ number of attractor basins.
Step 2: Distribute $S$ into $k$ attractor basins (nodes).
Step 3: Evaluate the distribution of examples in each attractor basin (node).
Step 4: If all the examples (S') of an attractor basin (node) belong to only one class, then label the attractor basin (leaf node) for that class.
Step 5: If examples (S') of an attractor basin belong to $K'$ number of classes, then **Partition** (S', $K'$).
Step 6: Stop.

## 4. Results

In this experiment 200 frames of video images with considerable variations in head poses, expressions, camera viewing angle are used. The 'face 94' color image database is used for still images. We selected 100 face images with considerable expression changes and minor variation in head turn, tilt and slant. The performance ratios are 90% for video image sequence and 92% for still images. The results of face detection for both still and video images are shown in figure 7 and figure 8. Figure 9 shows distinct connected components. Rotation transformation is shown in figure 10. Figure 11 shows the located features. Geometric parameters and DCT coefficients were given to a classification network for recognition.

## 5. Conclusion

In this paper, faces are detected and facial features are located from video and still images. 'NRC-IIT Facial video database' is used as image sequences and 'face 94 color image database' is used for still images. Skin pixels and non-skin pixels are separated and skin region identification is done by RGB color space. From the extracted skin region, skin pixels are grouped to some meaningful groups to identify the face region. From the face region, facial features are located using segmentation technique. Orientation correction is done by using eyes. Parameters like inter eye distance, nose length, mouth position, and DCT coefficients are computed which is used for a RBF based neural network. In this experiment only one image sequence is used for detection of faces.

Future work in the approach includes recognition of faces from video sequences. It is also proposed to analyze facial expressions from video image sequences.

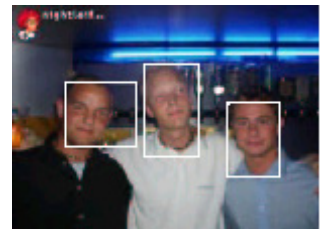

Figure 7. Face detection from still images

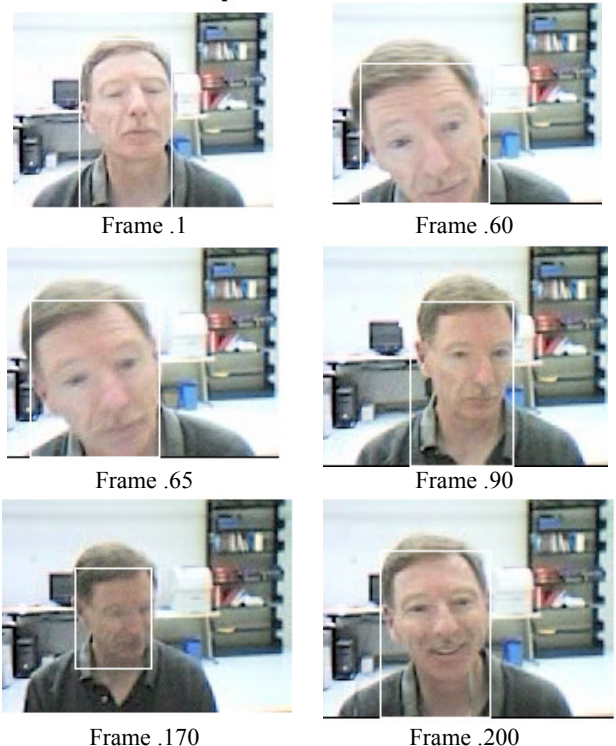

Frame .1                    Frame .60

Frame .65                   Frame .90

Frame .170                  Frame .200

Figure 8. Face detection from video images

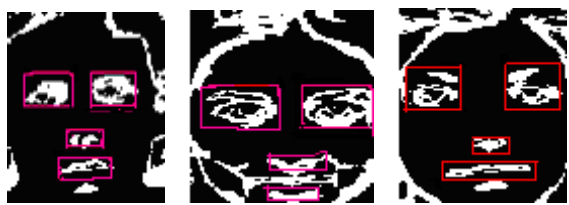

Figure 9. Distinct Connected Components

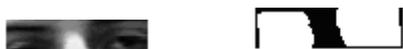

Figure 10.Eye region Image after angle Correction

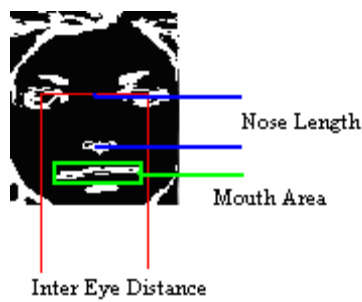

Figure 11. Located Features

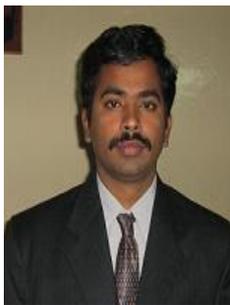

**P.KIRAN SREE** received his **B.Tech** in Computer Science & Engineering, from J.N.T.U and **M.E** in Computer Science & Engineering from Anna University. He has published many technical papers; both in international and national Journals .His areas of interests include Parallel Algorithms, Artificial Intelligence, Compile Design and Computer Networks. He also wrote books on Analysis of Algorithms, Theory of Computation and Artificial Intelligence. He was the reviewer for many IEEE Society Conferences in Artificial Intelligence and Networks. He was also member in many International Technical Committees. He is now associated with S.R.K Institute of Technology, Vijayawada.

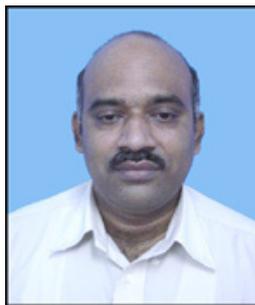

**Inampudi Ramesh Babu** received his **Ph.D** in Computer Science from Acharya Nagarjuna University, **M.E** in Computer Engineering from Andhra University, **B.E** in Electronics & Communication Engg from University of Mysore. He is currently working as Head & Professor in the department of computer science, Nagarjuna University. Also he is the senate member of the same University from 2006. His areas of interest are image processing & its applications, and he is currently supervising 10 Ph.D students who are working in different areas of image processing. He is the senior member of IEEE, and published 35 papers in international conferences and journals.